\documentclass[11pt,a4paper]{article}
\usepackage{amsmath,amssymb}
\usepackage{bm}
\usepackage[dvipdfmx]{graphicx}
\usepackage{float}
\usepackage{subfig}
\usepackage{authblk}
\usepackage{geometry}
\usepackage{hyperref}
\usepackage{color}
\usepackage{soul}
\usepackage[textsize=tiny]{todonotes}
\providecommand{\keywords}[1]{\textbf{\textit{Key words:}} #1}
\title{On the influence of Dice loss function in multi-class organ segmentation of abdominal CT using 3D fully convolutional networks}
\author[1]{Chen SHEN}
\author[1]{Holger R. ROTH}
\author[2]{Hirohisa ODA}
\author[1]{Masahiro ODA}
\author[1]{Yuichiro HAYASHI}
\author[3]{Kazunari MISAWA}
\author[1]{Kensaku MORI}
\affil[1]{Graduate School of Informatics, Nagoya University}
\affil[2]{Graduate School of Information Science, Nagoya University}
\affil[3]{Aichi Cancer Center Hospital}
\date{}
\begin{document}
\maketitle
\begin{abstract}
\noindent Deep learning-based methods achieved impressive results for the segmentation of medical images. With the development of 3D fully convolutional networks (FCNs), it has become feasible to produce improved results for multi-organ segmentation of 3D computed tomography (CT) images. The results of multi-organ segmentation using deep learning-based methods not only depend on the choice of networks architecture, but also strongly rely on the choice of loss function. In this paper, we present a discussion on the influence of Dice-based loss functions for multi-class organ segmentation using a dataset of abdominal CT volumes. We investigated three different types of weighting the Dice loss functions based on class label frequencies (uniform, simple and square) and evaluate their influence on segmentation accuracies. Furthermore, we compared the influence of different initial learning rates. We achieved average Dice scores of 81.3\%, 59.5\% and 31.7\% for uniform, simple and square types of weighting when the learning rate is 0.001, and 78.2\%, 81.0\% and 58.5\% for each weighting when the learning rate is 0.01. Our experiments indicated a strong relationship between class balancing weights and initial learning rate in training.
\end{abstract}
\section{Introduction}

Multi-class organ segmentation of abdominal computed tomography (CT) images is important for medical image analysis. Abdominal organs can have large individual differences due to the shape and size variations, which makes development of automated segmentation methods challenging. With the rapid development of medical image devices and the increase in routinely acquired images, fully automatic multi-organ segmentation of medical images becomes especially important. The segmentation results can be widely utilized in computer-aided diagnosis and computer-assisted surgery. Hence, much research \cite{1}\cite{2}\cite{3} has focused on automated organ segmentation from CT volumes. However, achieving high segmentation accuracy is always challenging. One of the reasons is the low contrast to surrounding tissues which makes segmentations difficult. Failure of segmentation results in a reduction of diagnostic quality. Therefore, improving the segmentation accuracy is an active area of research. Recently, deep learning-based methods achieved impressive segmentation results on medical images. For example, convolutional neural networks (CNNs) make it easier to train models on a large datasets. Especially, 3D fully convolutional networks (FCNs) improved the accuracy of multi-organ segmentation from CT volumes. It is well-known that network architecture influences the result of segmentation. Furthermore, the segmentation accuracy relies on the choice of loss function.
For this research, we developed a 3D FCN which can learn to automatically segment organs on CT volumes from a set of CT and labelled images, and discuss the influence of loss functions and initial learning rates. 
The paper is organized as the follows. In Section 2, we describe the methods we utilized in detail. Section 3 gives the experiments and results, and we provide a discussion in Section 4. 

\section{Methods}

\subsection{Overview}
FCNs have the ability to train models for segmentation from images in end-to-end fashion. With the development of FCNs, improved segmentation results have been reported\cite{4}. In this paper, we utilize a 3D U-Net architecture for multi-organ segmentation, similar to the one proposed by \c{C}i\c{c}ek et al\cite{5}. This network architecture is a type of 3D FCN designed for end-to-end image segmentation.
We investigated three different types of weighting models for the Dice loss functions in order to evaluate their performances and compared segmentation accuracy of each organ for the different situations. In addition, we performed training on the same dataset with different initial learning rates $\mu$, and compared how these parameters influence the segmentation results.
\subsection{3D fully convolutional network}
With the improvement of CNN architectures and GPU memory, we are able to train a larger number of annotated 3D medical CT volumes to improve the segmentation results. We utilize a training set of abdominal CT volumes and labels $S={(I_n,L_n),n=1,...,N}$, where $I_n$ represents CT volumes, $L_n$ represents the ground truth label volumes. We define n as the index of volumes and N is the number of training volumes.

In this work, we utilize a 3D U-Net type FCN with constant input size that is trained by randomly cropping subvolumes from the training data. Hence, we can obtain a trained model, which can segment the full 3D volume through subvolume prediction in testing. We chose an input size of 64×64×64 that allows to use mini-batch sizes of three subvolumes sampled from different training patients.
\subsection{Dice loss function}
The Dice similarity coefficient (DSC) measures the amount of agreement between two image regions\cite{6}. It is widely used as a metric to evaluate the segmentation performance with the given ground truth in medical images. The DSC is defined in (1), we utilize $|~~|$ to indicate the number of foreground voxels in the ground truth and segmentation images\cite{7}.
\begin{equation}
DSC=\frac{2|S\cap R|}{|S|+|R|}
\label{equ:dice_score}
\end{equation}
where $S$ is the segmentation result and $R$ is the corresponding ground truth label. This function however is not differentiable and hence cannot directly be used as a loss function in deep learning. Hence, continuous versions of the Dice score have been proposed that allow differentiation and can be used as loss in optimization schemes based on stochastic gradient descent\cite{8}:
\begin{equation}
\mathcal{L}_{DSC}=-\frac{2\sum_{i}^{N} s_ir_i}{\sum_{i}^{N} s_i+\sum_{i}^{N} r_i}
\label{equ:dice_loss}
\end{equation}
where $s_i$ and $r_i$ represent the continuous values of the \textit{softmax} prediction map $\in [0,\dots,1]$ and the ground truth at each voxel $i$, respectively. Using the formulation in \ref{equ:dice_loss}, we investigate three types of weighting based on class voxel frequencies in the whole training dataset. We defined three types of weighting factors $W_u$, $W_s$ and $W_q$ for \textit{uniform}, \textit{simple} and \textit{square} weighting. The equations are as follows:
\begin{equation}
W_u=1
\end{equation}
\begin{equation}
W_s=\frac{N}{L|R_l|+\epsilon}
\end{equation}
\begin{equation}
W_q=\frac{N}{L|R_l|^{2}+\epsilon}
\end{equation}
here, $L$ is the number of labels and $|R_l|$ is the number voxels in class $l$. We set $\epsilon=1$ in this experiment in order to avoid division by zero.\\
We calculate the multi-class Dice loss function using \ref{equ:multi_dice_loss}:
\begin{equation}
\mathcal{L}_{MD}=\frac{1}{L}\sum_{l}^{L} w_l  L_{DSC}
\label{equ:multi_dice_loss}
\end{equation}
with $w_l$ indicating the weight for class $l$, computed from one of the weighting types $W_u, W_s, W_q$, .
\begin{figure*}[t]
    \begin{center}
    \begin{tabular}{ccc}
        \subfloat[Initial learning rate $\mu=0.001,W_u$]{\includegraphics[height=3.2cm]{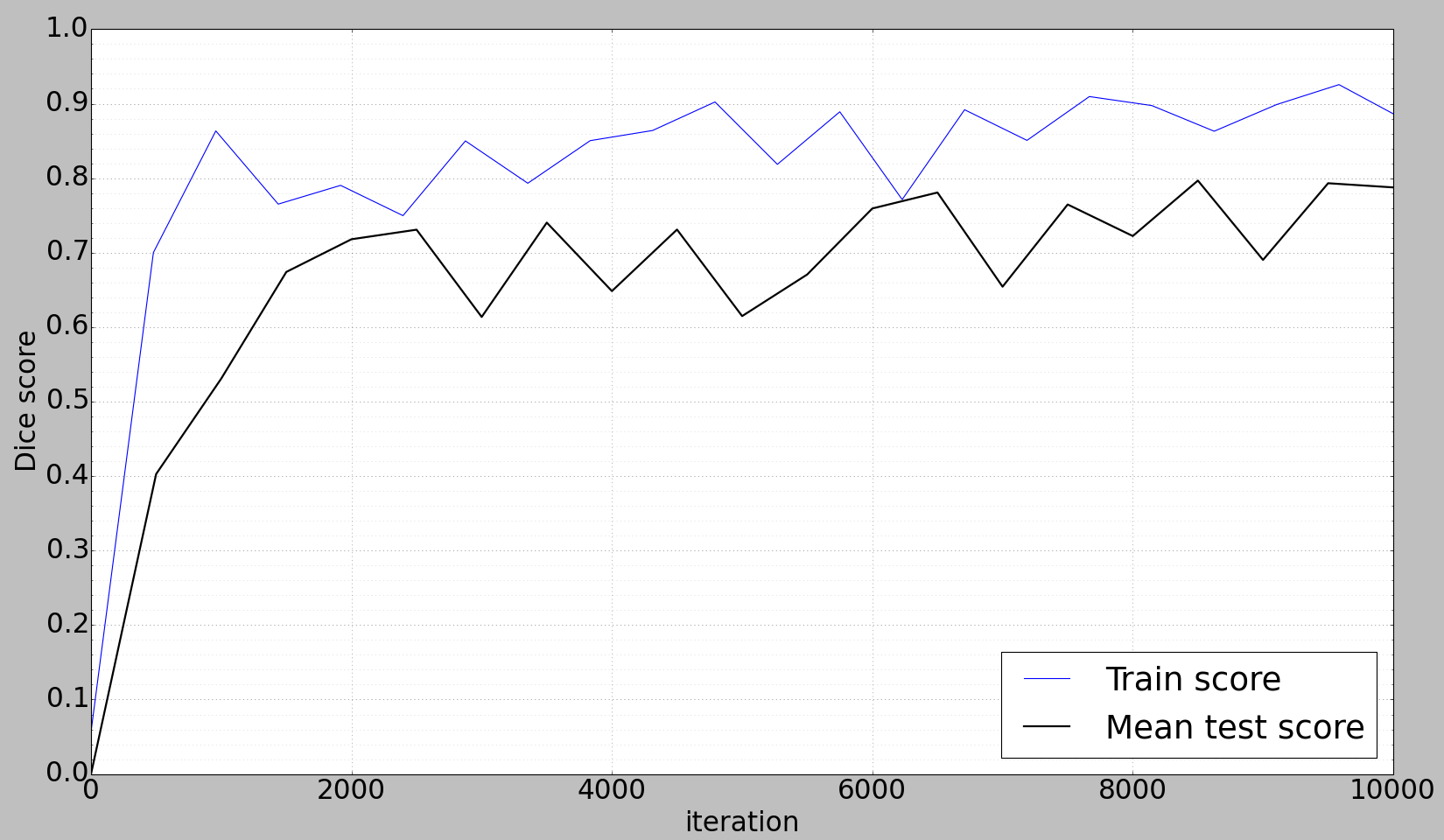}}
        \subfloat[Initial learning rate $\mu=0.001,W_s$]{\includegraphics[height=3.2cm]{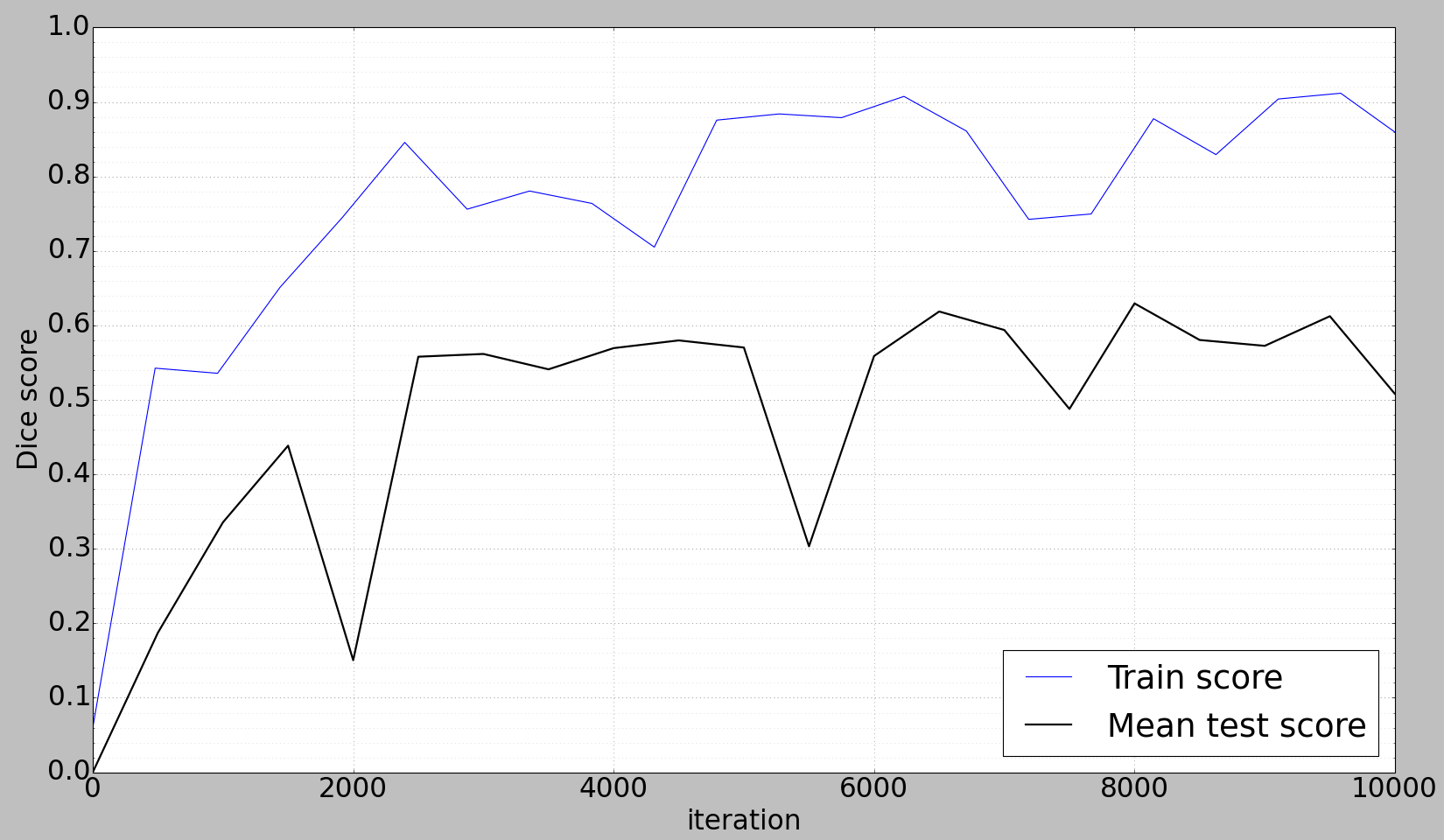}}
        \subfloat[Initial learning rate $\mu=0.001,W_q$]{\includegraphics[height=3.2cm]{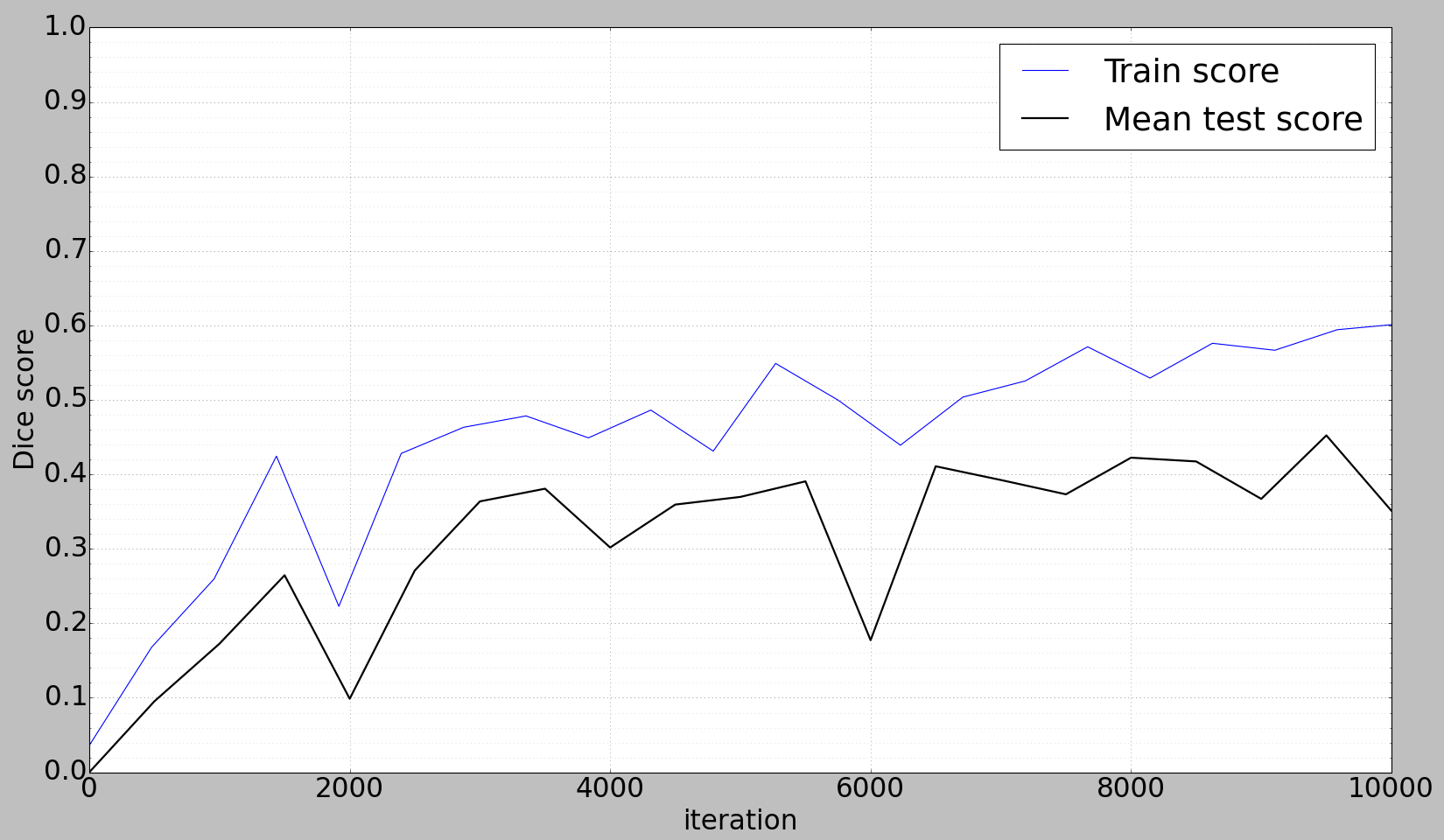}}\\
        \subfloat[Initial learning rate $\mu=0.01,W_u$]{\includegraphics[height=3.2cm]{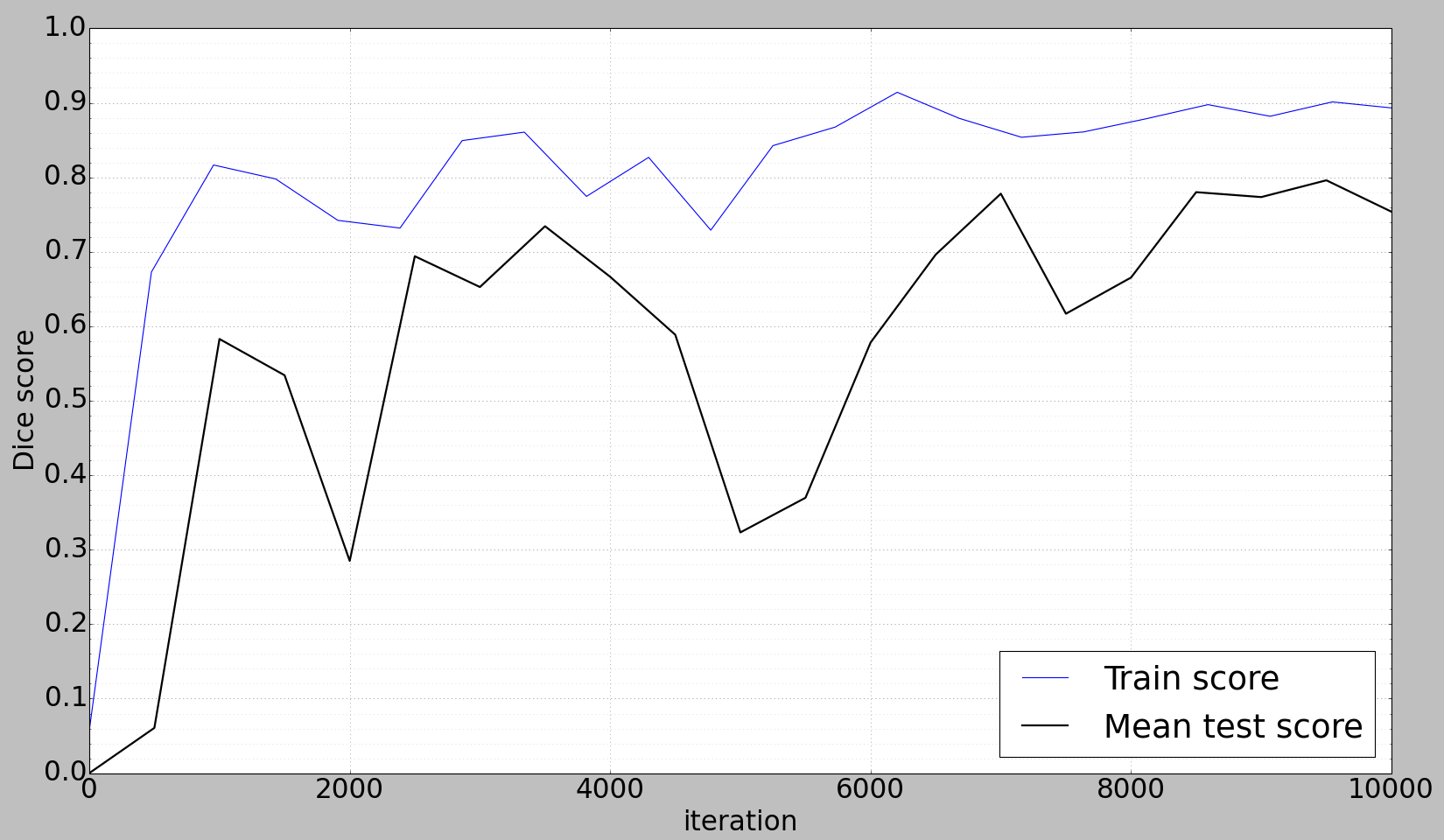}}
        \subfloat[Initial learning rate $\mu=0.01,W_s$]{\includegraphics[height=3.2cm]{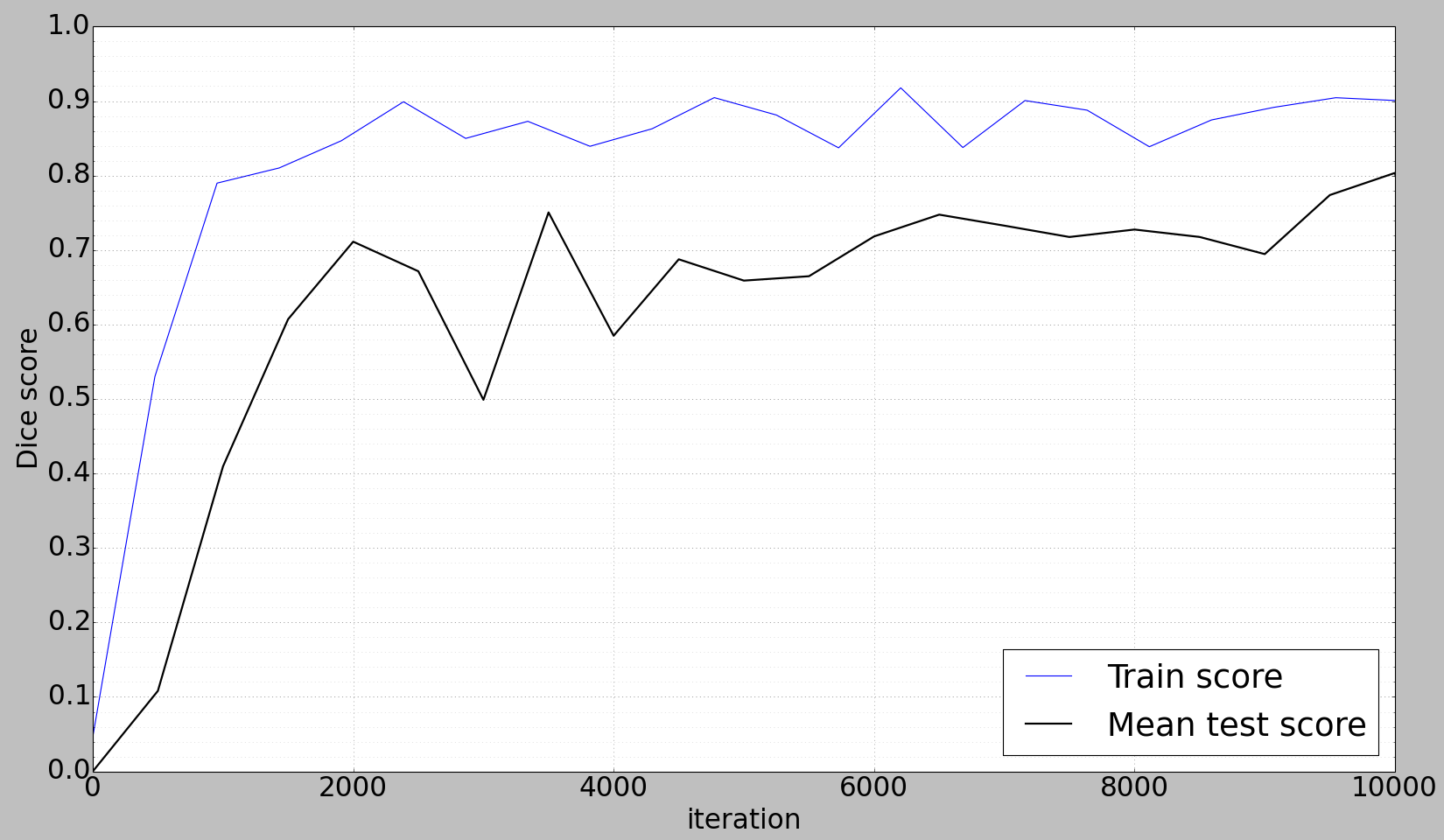}}
        \subfloat[Initial learning rate $\mu=0.01,W_q$]{\includegraphics[height=3.2cm]{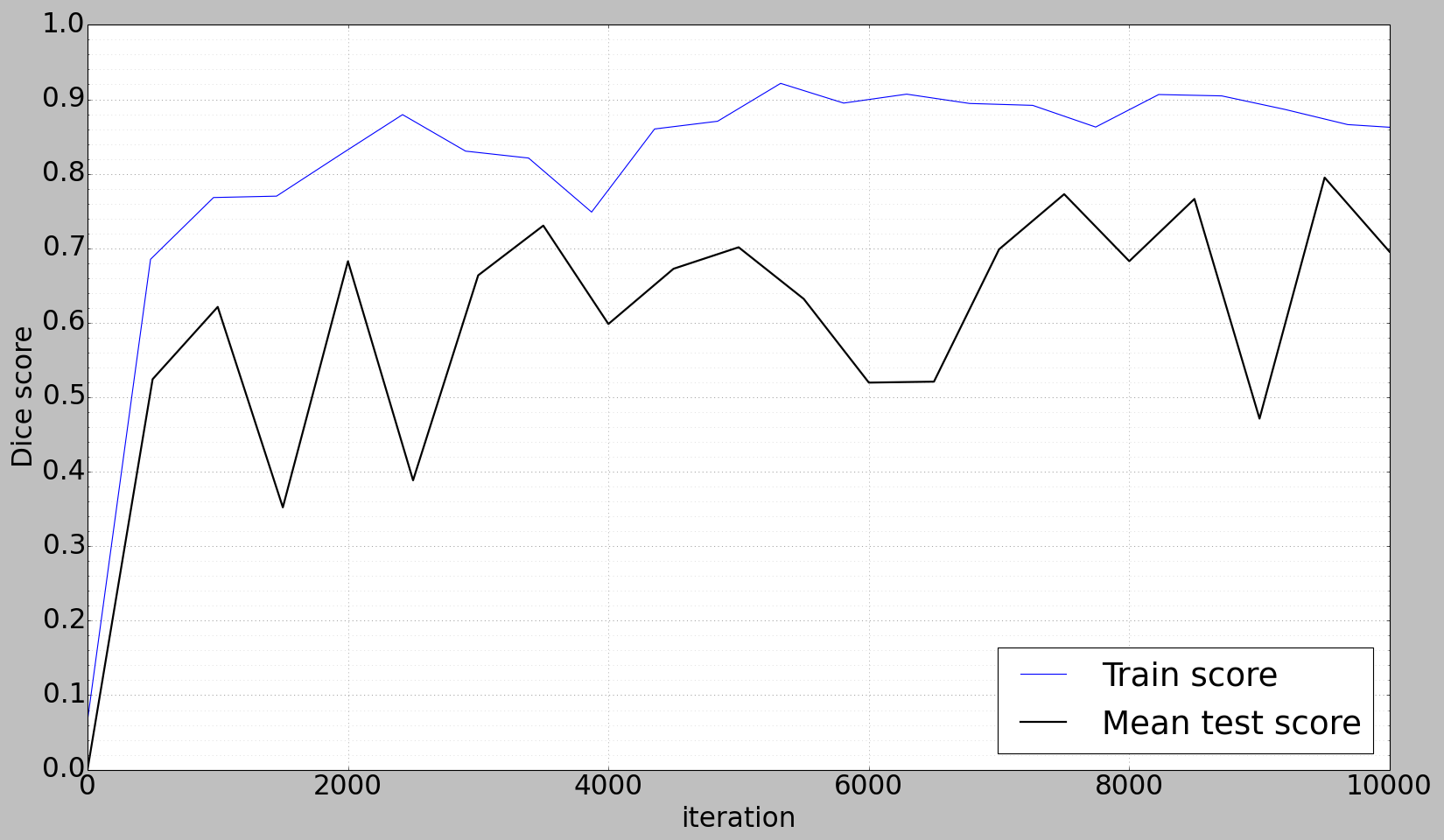}}\\
	\end{tabular}   
    \end{center} 
      \caption{Learning curves for different models.}
      \label{fig:learning_curves}
\end{figure*}
\begin{table}[tb]
\caption{Average Dice scores of different type of weighting schemes (\textit{uniform}, \textit{simple}, \textit{square}). $\mu_1$ and $\mu_2$ is the value of initial learning rate. We define $\mu_1$=0.001 and $\mu_2$=0.01 in this experiment. The highest accuracies for each organ are written in \textbf{bold} text.}
\label{tab:dice_scores}
\begin{center}
\scalebox{0.9}{
\begin{tabular}{c|c|c|c|c|c|c}
\hline
Dice type& $W_u,\mu_1=0.001$& $W_s,\mu_1=0.001$& $W_q,\mu_1=0.001$ & $W_u,\mu_2=0.01$& $W_s,\mu_2=0.01$ & $W_q,\mu_2=0.01$\\
\hline
background& \textbf{99.9\%} & 99.1\% & 79.2\% & 99.8\% &99.8\% &99.6\% \\
\hline
artery& 80.4\% &63.6\% &63.4\% &79.5\% &\textbf{89.2\%} &72.4\% \\
\hline
vein& 78.6\% &68.8\% &69.1\% &75.1\% &\textbf{79.3\%} &73.0\% \\
\hline
liver& \textbf{96.5\%} &73.4\% &10.1\% &95.6\% &96.3\% &87.7\% \\
\hline
spleen& \textbf{94.7\%} &71.6\% &0\% & 92.0\% &93.9\% &91.6\% \\
\hline
stomach& \textbf{96.3\%} & 64.1\% & 6.9\% & 93.9\% & 96.1\% & 82.4\%\\
\hline
gallbladder& 77.3\% & 74.7\% & 54.3\% &73.2\% & \textbf{80.8\%} & 54.2\% \\
\hline
pancreas& 82.7\% & 78.0\% & 69.1\% & 82.4\% & \textbf{84.7\%} & 70.3\% \\
\hline
AVG& 88.3\% & 74.2\% & 44.0\% & 86.4\%& \textbf{88.9\%} & 78.9\% \\
\hline
MAX& \textbf{99.9\%} & 99.1\% & 79.2\% & 99.8\% & 99.8\% & 99.6\%\\
\hline
MIN& 77.3\% & 63.6\% & 0.0\%& 73.2\% & \textbf{79.3\%} & 54.2\%\\
\hline
\end{tabular}}
\end{center}
\end{table}
\section{Experiments and results}
\begin{figure*}[htb]
\begin{center}
    \begin{tabular}{ccc}
        \subfloat[Ground truth of CT image]{\includegraphics[height=4.5cm]{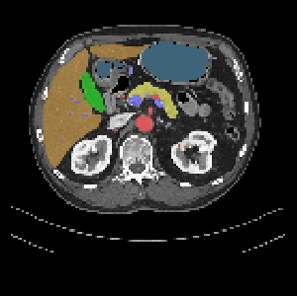}}&
        \subfloat[Learning rate $\mu=0.001$]{\includegraphics[height=4.5cm]{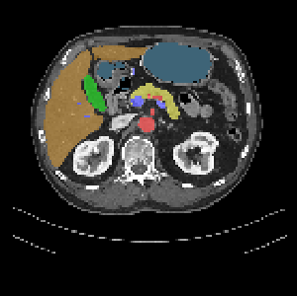}}&
        \subfloat[Learning rate $\mu=0.01$]{\includegraphics[height=4.5cm]{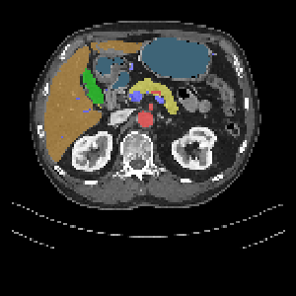}}\\
        \subfloat[Ground truth of 3D rendering]{\includegraphics[height=4.5cm]{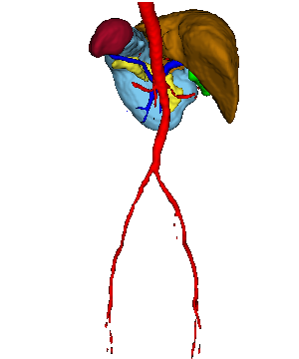}}&
        \subfloat[Learning rate $\mu=0.001$]{\includegraphics[height=4.5cm]{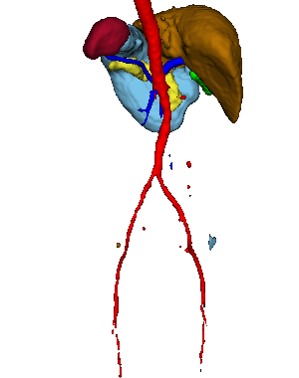}}&
        \subfloat[Learning rate $\mu=0.01$]{\includegraphics[height=4.5cm]{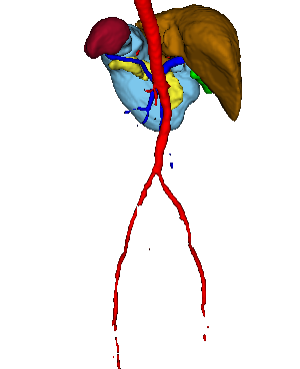}}\\
	\end{tabular}   
      \caption{Segmentation results for \textit{uniform} type weight. The images in top row are CT axial slices and in the bottom row are 3D surface renderings of the segmentation results.}
      \label{fig:seg_uniform}
\end{center}
\end{figure*}

\begin{figure*}[htp]
\begin{center}
    \begin{tabular}{ccc}       
        \subfloat[Ground truth of CT image]{\includegraphics[height=4.5cm]{gt_ct.png}}&
        \subfloat[Learning rate $\mu=0.001$]{\includegraphics[height=4.5cm]{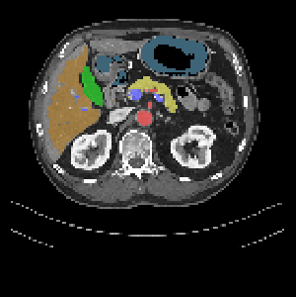}}&
        \subfloat[Learning rate $\mu=0.01$]{\includegraphics[height=4.5cm]{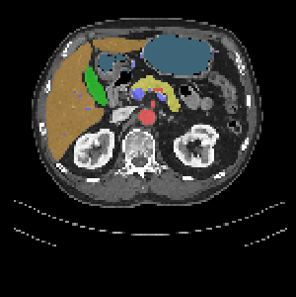}}\\
        \subfloat[Ground truth of 3D rendering]{\includegraphics[height=4.5cm]{gt.png}}&
        \subfloat[Learning rate $\mu=0.001$]{\includegraphics[height=4.5cm]{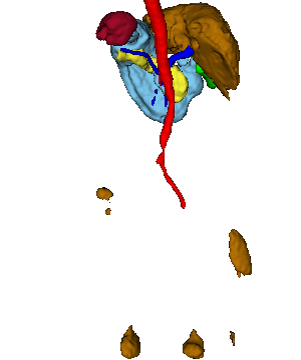}}&
        \subfloat[Learning rate $\mu=0.01$]{\includegraphics[height=4.5cm]{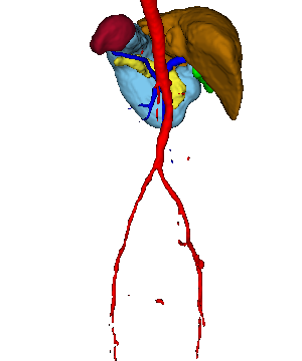}}\\
	\end{tabular}   
      \caption{Segmentation results for \textit{simple} type weight. The images in top row are CT axial slices and in the bottom row are 3D surface renderings of the segmentation results.}
      \label{fig:seg_simple}
\end{center}
\end{figure*}

\begin{figure*}[htp]
 \begin{center}
    \begin{tabular}{ccc}
        \subfloat[Ground truth of CT image]{\includegraphics[height=4.5cm]{gt_ct.png}}&
        \subfloat[Learning rate $\mu=0.001$]{\includegraphics[height=4.5cm]{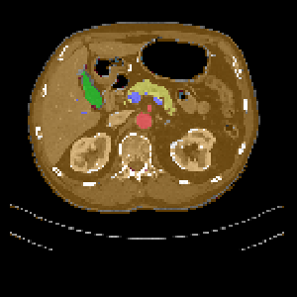}}&
        \subfloat[Learning rate $\mu=0.01$]{\includegraphics[height=4.5cm]{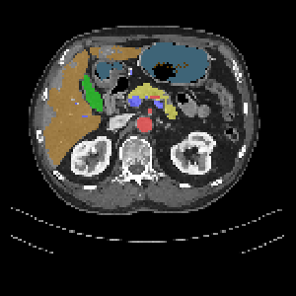}}\\
        \subfloat[Ground truth of 3D rendering]{\includegraphics[height=4.5cm]{gt.png}}&
        \subfloat[Learning rate $\mu=0.001$]{\includegraphics[height=4.5cm]{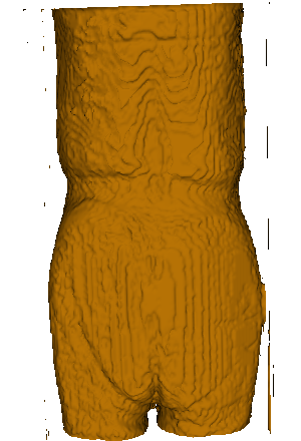}}&
        \subfloat[Learning rate $\mu=0.01$tab]{\includegraphics[height=4.5cm]{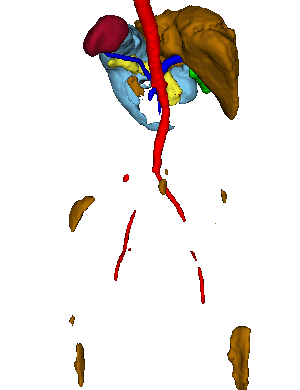}}\\
	\end{tabular}   
      \caption{Segmentation results for \textit{square} type weight. The images in top row are CT axial slices and in the bottom row are 3D surface renderings of the segmentation results.}
      \label{fig:seg_square}
 \end{center}
\end{figure*}

We used 377 abdominal clinical CT volumes in portal venous phase acquired as pre-operative scans for gastric surgery planning. Each CT volume contains 460-1177 slices of 512 $\times$ 512 pixels. We down-sampled the volumes with a factor of four in all experiments. We evaluated our models using a random subset of 340 training and 37 testing patients. The number of labelled abdominal organs is seven, which includes the artery, portal vein, liver, spleen, stomach, gallbladder and pancreas. Our network produced eight prediction maps as output (including the seven organ classes plus the background). Ground truth was established manually using semi-automated segmentation tools like graph-cuts and region growing using the Pluto software\cite{10}.\\
We implement our models in Keras\cite{11} using the TensorFlow backend. We compared the Dice similarity scores of seven organs and background of the three different weight types: \textit{uniform}, \textit{simple} and \textit{square}. Furthermore, we compared different initial learning rates of $\mu_1=0.001$ and $\mu_2=0.01$ using Adam\cite{9} as optimization method. Training of each model was performed for 10,000 iterations which takes about two days on a NVIDIA GeForce GTX 1080 GPU with 8GB memory. The segmentation results of the models with different weightings and learning rates are shown in Table \ref{tab:dice_scores}. The average Dice score performances were 88.3\% for \textit{uniform}, 74.2\% for \textit{simple} and 44.0\% for \textit{square} weighting　when the initial learning rate $\mu_1$=0.001. When the initial learning rate was $\mu_2$=0.01, the Dice score for \textit{uniform}, \textit{simple} and \textit{square} is 86.4\%, 88.9\% and 78.9\%, respectively.\\
Figure \ref{fig:learning_curves} depicts the learning curve for comparison of different weighting types. Figures \ref{fig:seg_uniform}, \ref{fig:seg_simple} and \ref{fig:seg_square} show the segmentation results of each weighting scheme.
\section{Discussion}
The results of our experiments indicate that the weighting of the Dice loss function and the initial learning rate both affected the performance of multi-organ segmentation in CT volumes. 
For the type of Dice loss function, the results showed clearly in Table \ref{tab:dice_scores} that when the learning rate was 0.001, the \textit{uniform} model performed the highest on average DSC of eight classes in the CT volumes. It achieved impressive result for liver, spleen and stomach with average DSC of 93.7\%, 91.6\% and 91.1\%, respectively. For other organs, although the results were not the best, they were still acceptable. The \textit{simple} and \textit{square} models performed much worse than \textit{uniform} on liver, spleen, and stomach segmentation. However, the \textit{simple} model performed best when the learning rate was increased to 0.01. The artery, portal vein, and pancreas performed best on this case, and the DSC of liver, spleen, stomach and gallbladder are also higher. As for the \textit{uniform} weights, it kept a stable accuracy.\\
For the learning rate, the results also can be seen from Table \ref{tab:dice_scores}. As we increased the learning rate from 0.001 to 0.01, the performances of \textit{simple} and \textit{square} models improved markedly. Especially for the \textit{simple} type, the average DSC was raised from 59.5\% to 81.0\%. Furthermore, the results on artery, portal vein, and pancreas showed the best with the \textit{simple} model when the initial learning rate was 0.01. Also, the \textit{square} model performed much better on segmenting the liver, spleen, and stomach. Therefore, we can conclude that a higher learning rate was beneficial for \textit{simple} and \textit{square} type models in multi-organ segmentation. We inferred that the DSC of the \textit{square} model could be even higher after the same number of iterations, if the learning rate was set to 0.1.\\
Moreover, the learning curves showed in Figure \ref{fig:learning_curves} indicated that the training converges to stable results for the \textit{uniform} weighting when the iteration is 10,000. However, for the \textit{simple} and \textit{square} weighting model, we can predict that the performance could continue to grow if training would be continued. We assume that a convergence would be achieved after 20,000 iterations.\\
Utilizing three types of weighting and two different learning rates, the relationship between weighting type and learning rate can be observed from these experiments. We assume that the iteration number will also influence the results, but might result in overfitting. Our experiments indicate that when introducing a class balancing weight, the initial learning rates and number of iterations have to be adjusted appropriately in order to achieve improvements in the segmentation accuracy.
\section{Conclusion}
We employed three types of weighting models for a Dice-based loss function and evaluated the segmentation accuracy for multiple organs with two different initial learning rates. These different types of weight models shows the influence on the multi-organ segmentation in CT volumes using FCN. The results depict that the class balancing weights and initial learning rates influence the performance of multi-organ segmentation in CT volumes. 

While we did not apply any data augmentation schemes, our results indicate no cases of strong overfitting, which points to a sufficiently large training dataset. Still, for future work, we can augment the original images and study how augmentation affects the segmentation accuracy. Moreover, we can match the best type of weighting and learning rate for single organs to achieve improvements in segmentation accuracy by combining the results from different models. With the availability of higher GPU memory and even larger datasets, the performance for automatic multi-organ segmentation is likely to increase.
\section{Acknowledgments}
This work was supported by MEXT KAKENHI (26108006, 26560255, 25242047, 17H00867, 15H01116) and the JPSP International Bilateral Collaboration Grant.

\end{document}